\documentclass[runningheads]{llncs}
\usepackage[T1]{fontenc}
\usepackage{graphicx}
\usepackage{hyperref}
\usepackage[final]{pdfpages}

\usepackage{color}

\usepackage[]{xcolor}

\def\showcomments{0}

\newcommand{\comment}[1]{\if\showcomments1\textbf{\textcolor{red}{#1}}\fi}
\newcommand{\commentSJ}[1]{\if\showcomments1\textbf{\textcolor{RawSienna}{[SJ: #1]}}\fi}
\newcommand{\commentRR}[1]{\if\showcomments1\textbf{\textcolor{JungleGreen}{[RR: #1]}}\fi}
\newcommand{\commentJD}[1]{\if\showcomments1\textbf{\textcolor{WildStrawberry}{[JD: #1]}}\fi}
\newcommand{\commentSG}[1]{\if\showcomments1\textbf{\textcolor{Plum}{[SG: #1]}}\fi}
\newcommand{\commentMW}[1]{\if\showcomments1\textbf{\textcolor{BurntOrange}{[MW: #1]}}\fi}
\newcommand{\commentPS}[1]{\if\showcomments1\textbf{\textcolor{NavyBlue}{[PS: #1]}}\fi}
\newcommand{\commentHH}[1]{\if\showcomments1\textbf{\textcolor{OliveGreen}{[HH: #1]}}\fi}
\newcommand{\deleteSuggestion}[1]{\if\showcomments1\textbf{\textcolor{Red}{[Delete: #1]}}\fi}

\usepackage{cleveref}
\usepackage[pdf]{pstricks}
\begin{document}
\title{A Combined Approach of Process Mining and Rule-based AI for Study Planning and Monitoring in Higher Education}
\titlerunning{Process Mining and Rule-based AI for Study Planning and Monitoring}
%
\author{Miriam Wagner\inst{1} \and
Hayyan Helal\inst{2}\and
Rene Roepke\inst{3}\and
Sven Judel\inst{3}\and
Jens Doveren\inst{3}\and
Sergej Goerzen\inst{3}\and
Pouya Soudmand\inst{1} \and
Gerhard Lakemeyer\inst{2}\and
Ulrik Schroeder\inst{3}\and
Wil van der Aalst\inst{1}}

%
\authorrunning{M. Wagner et al.}
%
\institute{Process and Data Science (PADS), RWTH Aachen University, Germany
\email{\{wagner;pouya.soudmand;wvdaalst\}@pads.rwth-aachen.de}\\
\and
Knowledge-Based Systems Group, RWTH Aachen University, Germany
\email{\{helal;gerhard\}@kbsg.rwth-aachen.de}\\
\and
Learning Technologies Research Group, RWTH Aachen University, Germany\\
\email{\{roepke;judel;doveren;goerzen\}@cs.rwth-aachen.de}\\
}
\maketitle              
\begin{abstract}
This paper presents an approach of using methods of process mining and rule-based artificial intelligence to analyze and understand study paths of students based on campus management system data and study program models.
Process mining techniques are used to characterize successful study paths, as well as to detect and visualize deviations from expected plans.
These insights are combined with recommendations and requirements of the corresponding study programs extracted from examination regulations.
Here, event calculus and answer set programming are used to provide models of the study programs which support planning and conformance checking while providing feedback on possible study plan violations.
In its combination, process mining and rule-based artificial intelligence are used to support study planning and monitoring by deriving rules and recommendations for guiding students to more suitable study paths with higher success rates.
Two applications will be implemented, one for students and one for study program designers.
\end{abstract}
%

\section{Introduction}
\comment{Since not all potential readers (myself included) are necessarily familiar with the course grading system used in higher education institutions in Germany, I advise the authors to add a short description of such a system: course registration, exam registration and results, the relationship between exams and course grading, course final grades, among others.
-> ignore?}
In higher education, study programs usually come with an idealized, recommended study plan.
However, given how students have different capacities to study due to circumstances like part-time jobs or child care, and how one deviation from the intended study plan can have ripple effects spanning several semesters, in reality, a large number of different study paths can be observed.
Further, capacity limits like room sizes or the amount of supervision that lecturers can provide make the planning of study paths more complex.
Even though individualized study paths are possible due to the flexibility in study programs and their curriculum, students may need assistance and guidance in planning their studies.
Software systems that assist students and study program designers in planning might do so by analyzing the large amounts of data in higher education institutions~\cite{daniel.2015}.
Of particular interest in this context are event data extracted from \textit{Campus Management Systems} (CMS) including course enrollments, exam registrations and grade entries.
To this purpose the \textit{AIStudyBuddy} project - a cooperation between RWTH Aachen University, Ruhr University Bochum and University of Wuppertal - is set up.
For preliminary analyses, we received access to the CMS data of two Bachelor programs, Computer Science and Mechanical Engineering, at RWTH Aachen University.
Within the project, it will be investigated how to preprocess the data of all partners to apply the preliminary as well as the further intended analyses.

\begin{figure}[h]
\includegraphics[width=0.9\textwidth]{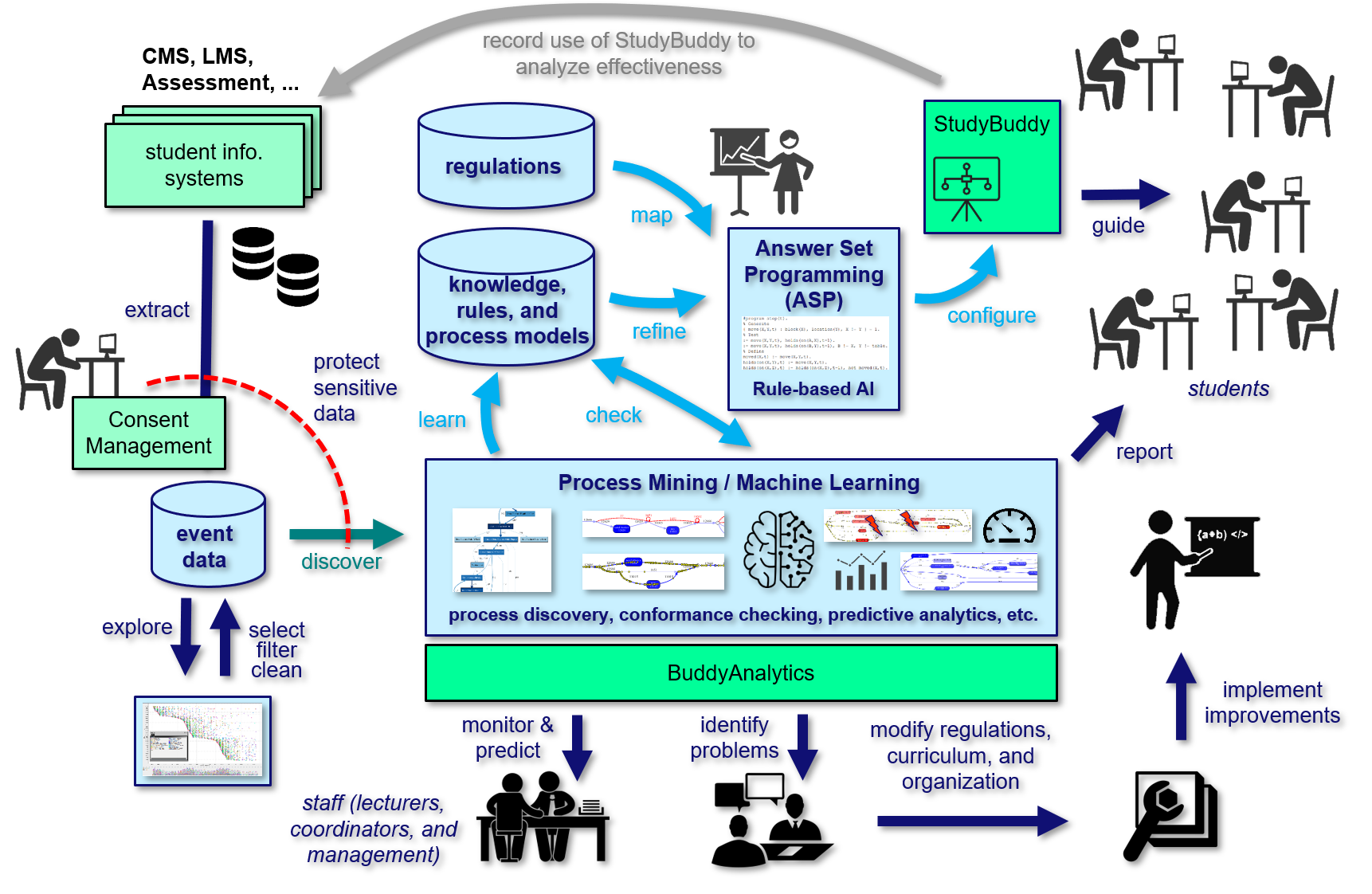}
\caption{Overview of the project, showing the two parts: \textit{StudyBuddy} and \textit{BuddyAnalytics} and their relationships to the different systems and techniques.} \label{fig:overview}
\end{figure}

The aim of the project is to develop two applications: an intelligent planning tool for students and an analytics dashboard for study program designers (see \autoref{fig:overview}).
Both will be powered by a combination of rule-based \textit{Artificial Intelligence} (AI) and \textit{Process Mining} (PM) approaches.
The implementation and evaluation of this combination's ability to efficiently generate rich feedback when checking the conformance to formal study plans is a key aspect of this project.
This feedback will include PM results in the form of \textit{recommendations}, which do not result from explicit regulations but rather historic study path data.

The planning tool for students, \textit{StudyBuddy}, will use rule-based AI to check preliminary plans against an encoding of study program regulations.
It will be able to provide immediate high-quality feedback regarding any potential conflicts in students' study plans.
In addition to the rules explicitly codified in institutional regulations, the tool will have a notion of recommendations, which result from analyzing historical CMS data using PM approaches and finding characterizations of successful 
paths, e.g., finished in standard period of study.

The analytics dashboard, \textit{BuddyAnalytics}, will enable study program designers to explore the PM results for the process of \textit{Curriculum Analytics}.
Process models of recommended study plans can be compared to study paths in the data to detect possible deviations or favorable routes.
Various study path analyses could support monitoring and help study program designers as well as student counseling services to support successful study paths and intervene in misguided study planning by providing individualized plans.

The paper is structured as follows:
\Cref{sec:rw} presents relevant related work in the fields of PM, rule-based AI and curriculum analytics.
\Cref{sec:approach} introduces the aim of addressing individualized study planning for students and data-driven study monitoring for study program designers in a combined approach.
The current state as well as challenges of the project are described in \Cref{sec:state}, while \Cref{sec:fw} presents objectives of future work.
\Cref{sec:conclusion} concludes the paper.

\section{Related Work}\label{sec:rw}
\subsection{Process Mining in Education}\label{sec:rw:pm}
\comment{Salazar-Fernandez, J. P., Sepúlveda, M., Munoz-Gama, J., \& Nussbaum, M. (2021). Curricular analytics to characterize educational trajectories in high-failure rate courses that lead to late dropout. Applied Sciences, 11(4), 1436.
Salazar-Fernandez, J. P., Munoz-Gama, J., Maldonado-Mahauad, J., Bustamante, D., \& Sepúlveda, M. (2021). Backpack Process Model (BPPM): A Process Mining Approach for Curricular Analytics. Applied Sciences, 11(9), 4265.
-> Done
}
\comment{Typo on page 3, section 2.1 "is a of PM" < there seems to be a word missing 5 => DONE}

\textit{Educational Process Mining} (EPM) \cite{Bogarin.2018,Sypsas.2022} is a 
sub-field of PM \cite{vanderAalst.2018}, using various, commonly known PM techniques in the educational context, e.g. higher education.
While we focus on CMS data, most work in EPM has been done using \textit{Learning Management Systems} (LMS) data with similar aims.
In ~\cite{Pechenizkiy.2009}, two online exams have been analyzed using dotted chart analysis and process discovery with various miners. 
The applicability of standard methods provided in ProM in the context of LMS data is shown. 
In ~\cite{Bogarin.2018.2}, course-related student data has been extracted to visualize the learning processes using an inductive miner to help preventing failing the course.
``Uncover relationships between usage behavior and students' grades'' is the aim of \cite{Etinger.2020} by using \textit{Directly-Follow Graph} (DFG).
In~\cite{Cenka.2022}, a case study is described in which the LMS data of one course is analyzed using among other things DFG.
Also, in~\cite{Mathrani.2022}, data from an LMS is used and the creation of the event log is described in detail.
Those event logs are used for the creation of DFG.

Analyses of LMS data show that the PM techniques can be used in the educational context but while concentrating on the behavior of students in one course, \textit{Curriculum Mining} analyzes the different study paths a student can take~\cite{Pechenizkiy.2012} which is a substantial aspect in our work.
Here, different approaches exist:~\cite{Wang.2015,Schulte.2017} describe ways to use curriculum data to uncover the de-facto paths students take to, in the next step, recommend suitable follow-up courses.
To our knowledge, this next step has not been done.
~\cite{BuckEmden.2018} focuses on the study counselor perspective and uses, e.g., Fuzzy Miner and Inductive Visual Miner, to visualize the de-facto study paths and use those insights to improve the curriculum.
In~\cite{SalazarFernandez.2021}, the influence of failing a course on the study success is analyzed using mainly DFGs, while in~\cite{SalazarFernandez.2021b}, the analysis is done by modeling how students retake courses and the influence on study program dropouts.

Further, we will explore the application of conformance checking \cite{Carmona.2018}. Therefore, similar approaches to ours are reviewed.
An extended approach to conformance checking is multi-perspective conformance checking as in ~\cite{Mannhardt.2016}.
For our purpose, one reason to not extend this technique is that the Petri nets representing the study plan are hard to read when including all allowed behavior.
For example, allowing a course to be in different semesters might lead to repositioning other courses as well.
Furthermore, some rules that need to be represented are not connected to the model itself, e.g., credit point thresholds belonging to a semester and not to a course.
Those could be modeled using invisible transitions, which makes the model more complicated and less intuitive. 
\subsection{Related Work on Rule-based AI}\label{sec:rw:ai}
The goal of rule-based AI is to model the examination regulations and the module catalog in a machine-readable language that allows for dealing with and planning events.
For such scenarios, the combination of \textit{Answer Set Programming} (ASP)
and \textit{Event Calculus} (EC) is applied.
Both are based on a wider concept called \textit{non-monotonic reasoning}, which differentiates from \textit{monotonic reasoning} by the ability to retract already made implications based on further evidence \cite{brewka1997nonmonotonic}.

Non-monotonic reasoning can model \textit{defaults} as described in \cite{reiter1980logic}. Defaults are assumed to hold, but do not necessarily have to. For instance, \textit{Students typically take course X after they do course Y} will be modeled as a default, as it is a recommendation, not a requirement.
As long as the student does not plan anything against it, it will be considered in their planning.
Else, it will be ignored. 
A requirement on the other hand must be valid for all plans.

Looking for similar approaches, in \cite{banbara2019varvec}, the problem of curriculum-based course timetabling was solved using ASP, however using a mechanism other than EC.
While we consider recommendations to be defaults that must be typically followed, they should only ever result in a warning to the student, still giving the freedom to be deviated from.
In \cite{banbara2019varvec}, recommendations come in handy for planning, where the number of violations on them should be minimized.
Furthermore, the timetabling problem focuses much more on the availability requirement for courses rather than also considering the results (e.g. success or failure, \textit{Credit Points} (CPs) collected, ...) of these courses, which is the main focal point for us.

More generally, \textit{Declarative Conformance Checking} \cite{Carmona.2018} is a common application of rule-based AI to process models.
In \cite{Leoni.2012,Burattin.2016}, declarative rules are used instead of classical conformance checking based on Petri nets.
While \cite{Leoni.2012} just covers the activities for constraints, \cite{Burattin.2016} extended it with a data- and time-perspective.
Furthermore, \cite{baldoni.2011} has a wider model for requirements.
It specifies three kinds of requirements, which refer to the relation in time between events, e.g. an event has a succession requirement if there is an event that must be done in the future after doing it.
All three approaches use Linear Temporal Logic instead of ASP and EC, as it it suitable for modeling the three mentioned requirements.
For our purposes though, it makes the modeling of the contribution of an event to a specific result (e.g., CPs) harder, because our approach does not focus on the relation in time between events as much as the contributions of these events.

\subsection{Curriculum Analytics and Planning}\label{sec:rw:ca}
Having emerged as a sub-field of Learning Analytics, curriculum analytics aims to use educational data to drive evidence-based curriculum design and study program improvements \cite{hilliger_lessons_2022}. 
Leveraging the data gathered in educational institutions, it can help identify student's needs and reduce dropout rates \cite{daniel.2015}.
As such, different approaches and tools (e.g., \cite{Bendatu.2015,brown_taken_2018,heileman_characterizing_2017,Priyambada.2017}) have been developed to support the analysis of CMS or LMS data with the aim of helping instructors and program coordinators reflect on the curriculum and teaching practices.
While various data and PM approaches have been used to analyze study paths provided through CMS event data \cite{Bendatu.2015,Priyambada.2017}, curriculum sequencing and study planning was explored using semantic web concepts applied on examination regulations, with the overall aim of supporting curriculum authoring, i.e., the design of personalized curricula fulfilling a set of constraints \cite{baldoni.2011}.
Other approaches include recommender systems \cite{wong_sequence_2018} or genetic algorithms \cite{srisamutr_course_2018} to support students in course selection processes and fulfilling requirements of a study program.
To the best of our knowledge, however, no joint approach of PM and rule-based AI has yet been explored in order to support study planning and monitoring for students and study program designers.

\section{Approach}\label{sec:approach}
The aim of AIStudyBuddy is to support individualized study planning (for students) and monitoring (for study program designers).
\textit{Study planning} describes the students' activities of planning and scheduling modules, courses and exams throughout the complete course of a study program.
The examination regulations provide recommendations and requirements to describe a study program and the conditions for students to successfully earn a degree.
These may include a sample study plan recommending when to take which module or course and attempting to distribute CPs evenly over the standard period of study.
Students choose from the module catalog, a list of mandatory and elective modules.

While most students may start with the same recommended plan in their first semesters, deviations due to various reasons can occur at any time, e.g., working part-time may result in a reduced course load and delaying courses to the next year, thus, changing the complete plan and its duration.
Therefore, support for individualized planning as well as recommendations of suitable study paths are needed.
Further, the diversity of study paths and deviations from recommended study plans raises questions of how different students move through a study program, if certain modules or courses cause delays in the study plan, or whether a study program may need revisions.
Here, \textit{study monitoring} can be provided by analyzing students' traces in various systems used in the university.
In 
our project, we will initially focus on CMS data 
 and might include LMS data later.

In order to support students and study program designers in their respective tasks, a modular infrastructure (see \autoref{fig:overview}) with two primary applications for the target groups will be implemented. 
The application \textit{StudyBuddy} presents a web interface to guide and engage students in study planning activities.
As in many programs students do not necessarily have to follow a recommended plan and in later phases not even have recommendations.
To help finding suitable courses historic data can be used to give hints which combinations have been successful.
Furthermore, course-content is not always independent from other courses and a specific order might help to pass with higher chance.
It offers an overview of a student's study program and allows for creation and validation of individual study plans.
ASP and EC are used to model these regulations.
\comment{Another important question that remains undiscussed is how much useful information one can derive from historical data of students enrolments. One of the explicit goals is to offer suggestions to students about which path to follow, but there is no discussion of the value of offering suggestions based on historical data from other students.
> Do you assume that the order of courses in one's higher education carrier is one (main) driver for their success? -> präsentation: one driver that we can influence
> Do you assume that when an order works for one student, the same order should also work for another student? -> addressed by talking about combinations
> Do you assume that the preferred order of courses is constant over time (given that content, teachers, etc. are not constant)? -> content is more or less constant 
> Cfr. "These recommendations are the result of mining historic data of previous study paths for those with high success rates" (p. 6) discovery and recommendation (StudyBuddy)
}
Given a study plan, they can be used to generate feedback regarding violations and give recommendations.
These recommendations are the result of mining historic data of previous study paths for those with high success rates.

For study program designers, the application \textit{BuddyAnalytics} presents an interactive, web-based dashboard visualizing PM data analysis results.
Different methods, i.e., process discovery and conformance checking, can help to understand how different student cohorts behave throughout the course of the study program and identify deviations from recommended study plans.
Based on different indicators and questions by study program designers, student cohorts can be analyzed and insights into their paths can be gained.
Study program designers can evaluate and compare different study paths and further develop new redesigned study plans in an evidence-based way.

\section{Current State \& Challenges}\label{sec:state}
\comment{it is not completely clear to me what the contribution exactly is and what the state of implementation of the proposed tool is. 
Particularly, I wonder how many of the addressed concerns/solutions exceed the capabilities of simple SQL queries on CMSs? -> Add sentence about the simple numbers
Especially when checking for violations, it would be interesting to see some examples that justify difficult algorithms (because of the difficult logic in the rules) relates to conformance checking (BuddyAnalytics)
-> presentation example? @Hayyan
}
\comment{In the description of the work they have done, it would be interesting if the authors:
. show some conceptual examples of the event logs they have built. -> we have a description?
. explain how they will deal with the fact that several courses are taken in parallel in the same semester, which is difficult to represent in DFG.-> using Petri nets
. explain how they will deal with the different granularities (at the course level vs. at the exam level) that they plan to use. ->ignore
. show some of the rules they obtained. ->presentation?
}

\begin{figure}
\centering
\includegraphics[width=\textwidth]{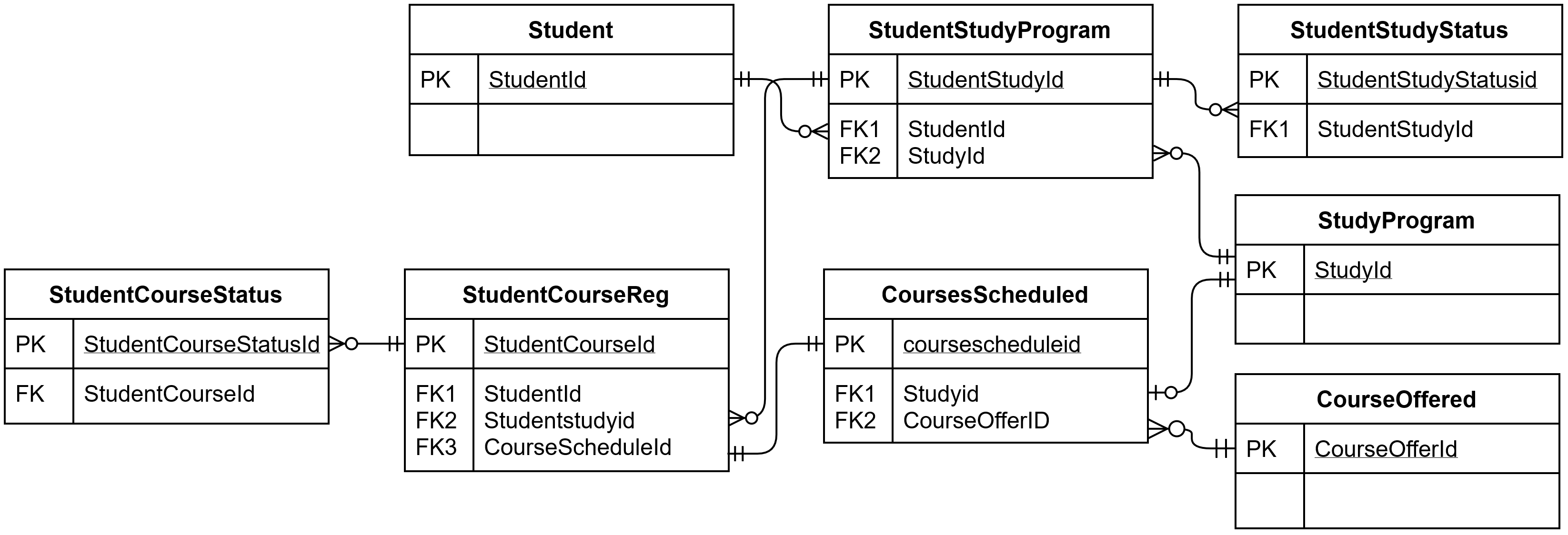}
\caption{A basic and generic data model for CMS data} \label{fig:model}
\end{figure}

The main data source for this project is the CMS of a university, which contains information about the students, courses, exams and their combination.
Later, the possibility to integrate LMS data will be explored.
As the project aims to be independent from the systems and study programs at different universities, a general data model has been created (see \autoref{fig:model}).
This model is the starting point for our project work and shows the general relation between courses and students as well as study programs.
The diagram does not include all possible data fields as they differ depending on the available data of a university.

Students can have multiple study programs, e.g., first do a Bachelor in Computer Science followed by a Master.
Each semester a student has a study status, e.g., \textit{enrolled} or \textit{semester on leave}.
The same offered course is scheduled in different semesters, e.g., \textit{Programming} is offered every winter semester, and in different study programs, e.g., \textit{Introduction to Data Science} is mandatory for a Master in Data Science but elective for a Master in Computer Science.
Students also have a status for scheduled courses during their study program, e.g., \textit{course passed}.

Until now, we explored data on exam information (ie., registrations and results).
The analyzed data includes Bachelor and Master Computer Science students as well as Mechanical Engineering Bachelor of RWTH Aachen University. 
Some standard KPIs used in various departments of universities that give meaningful insights about students, study programs or cohorts are:
\begin{itemize}
    \item Success rate of a course [in specific semesters] [for a cohort]
    \item Number of attempts a course is taken [on average] [for a cohort]
    \item Exams taken/passed in a [specific] semester [on average] [of a cohort]
    \item Average study duration [of a cohort]
    \item Percentage of dropouts [of a cohort] [in a predefined period]
\end{itemize}

A cohort can be defined based on the semester a group of students started, e.g., \textit{cohort WS21} refers to all students that started in winter semester 2021/2022.
It can also be defined by the amount of semesters students already studied or the examination regulations they belong to.
Different cohort definitions are needed to answer various questions about the behavior of students.

\begin{figure}
\centering
\includegraphics[width=\textwidth]{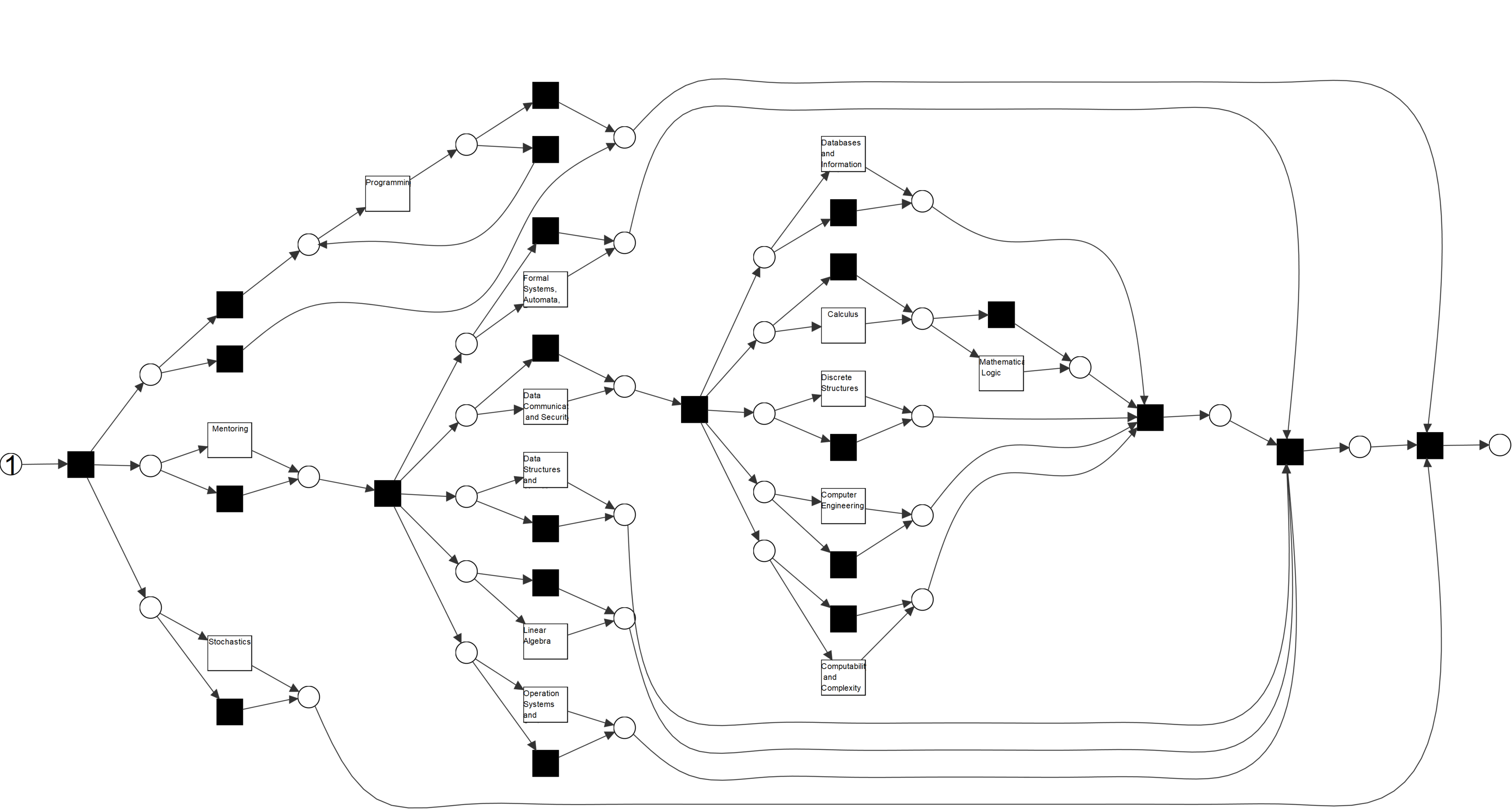}
\caption{Model created by ProM plugin "Mine Petri net with Inductive Miner" for data of students studying in examination regulation 2018 just using their mandatory courses}
\label{fig:disc}
\end{figure}

For more insights exceeding simple SQL queries used for descriptive statistics, the data is transferred into specific event logs, in which activities can be based just on courses and exams, or may even include additional information.
\comment{In the description of the work they have done, it would be interesting if the authors:
. show some conceptual examples of the event logs they have built. -> we have a description? what is meant with conceptual}
First, we concentrated on events describing the final status of exams for students.
A student can have multiple occurrences of a course, e.g. when they do not pass the exam in the first try or when they registered first, but in the end, they did not take it.
As a timestamp, the semester or the exact exam date can be used.
Also, some activities may have specific status dates, e.g., the date of the (de-)registration.
Those event logs can be used to create de-facto models showing the actual behavior of a group of students.
As 
model we use DFG, BPMN models, process trees and Petri nets, as shown in \Cref{fig:disc}, because they are easy to read also for non-specialists in PM.
\comment{The Petri net shown in Fig. 3 is not entirely meaningful; for example, all courses can be taken or skipped. -> Above and more in presentation?}

\begin{figure}
\centering
\includegraphics[width=\textwidth]{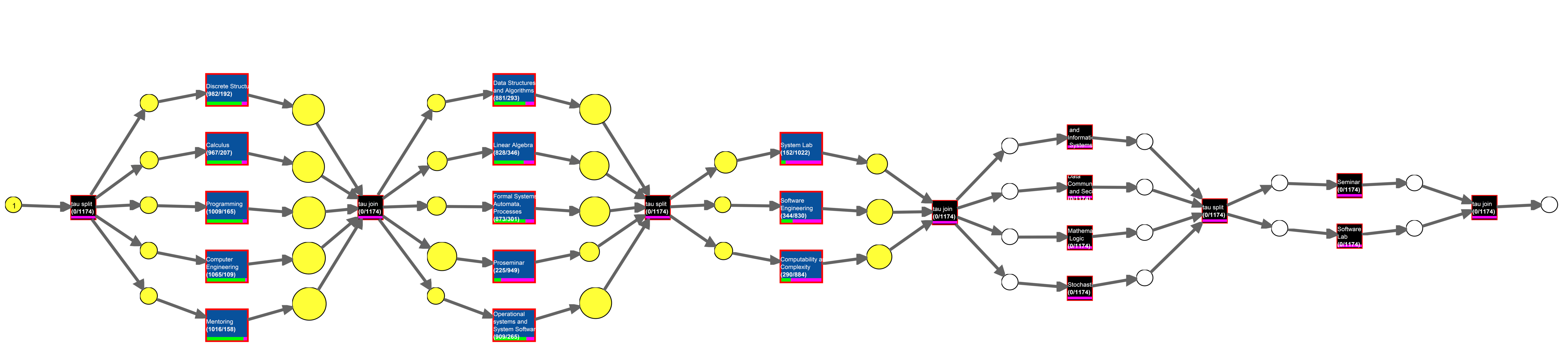}
\caption{Conformance checking result using ProM plugin "Replay a Log on Petri Net for Conformance Analysis" on data of students studying in examination regulation 2018 and a simplified Petri net model of the regulation}
\label{fig:conf}
\end{figure}

For useful insights, the multiple occurrence and the partial order of courses must be treated.
The partial order is caused by using, e.g., the scheduled semester, instead of the arbitrarily set exam dates, based on among others room capacities.
We tried out different solutions with the setting depending on the underlying questions that should be answered by the resulting model, e.g., when using a combination of exam attempt and course ID as the activity, the resulting de-facto model shows how courses are retried and visualizes better the real workload per semester.
In \Cref{fig:disc}, just the first occurrence of the course is used and all exams of a semester have the same date. 
Semester-blocks are visible, especially when the offered semester of a course is in mind, e.g., \textit{Programming} and \textit{Calculus} are offered in the winter semester.
The examination regulation of 2018 states that it should be taken in the first semester.
Compared to the (simplified) recommended plan (see \Cref{fig:conf}) \textit{Mentoring} occurs two semesters before \textit{Calculus}, while they should be concurrent.
\textit{Data Communication and Security} is taken two semesters earlier than planned and before courses that should precede it, e.g., \textit{Computability and Complexity}.
Those models give a first impression of the actual study path but need interpretation.

As a simpler approach to the later proposed combination of ASP and classical 
PM conformance checking, we explored the possibility of creating de-jure models based on the recommended study plan.
We used Petri nets since they can cover course concurrency and are still understandable by non-experts.
The de-jure model in \Cref{fig:conf} shows the main recommended path. Note, the data was just available including the third semester and later courses are invisible.
Using Petri nets and 
conformance checking this recommendation becomes a requirement. 
\comment{If the only result we can obtain from conformance checking is which courses are taken in the semester they are supposed to, I do not think we are taking full advantage of those algorithms. Applying a simple query to the database will provide the same results. -> Future work}

The results of classical conformance checking are still useful to find typical deviation points, e.g., \textit{Linear Algebra} tends to be taken in a different semester than proposed.
Also, when filtering on the first exam attempts, the resulting insights are different from filtering on the successful passing of exams.
Filtered on the first attempt, we can see how many students actually tried to follow the plan, while filtered on the passed exams indicates the success route.
When we have a high percentage of students that try to follow the recommended study plan, but just a low percentage that succeeds, this may be a warning for study program designers that the rules may need to be adapted to improve the recommendation and thereby increase the study success of students.

Our findings show that in later semesters, students deviate more from the recommended study plan, which can be explained by delays occurring earlier in their study.
What is not modeled by the Petri net here is that for \textit{Seminar} (semester 5), \textit{Proseminar} (semester 2) is a prerequisite.
Therefore, \textit{Proseminar} has to be taken before \textit{Seminar} and forms a \textit{requirement}.
Including those additional requirements and all already planned exceptions from the original plan, those models are fast becoming so called spaghetti models and lose a lot of their readability.
Lastly, additional constraints, e.g., credit point constraints such as \textit{at the end of the third semester, at least 60 CPs must have been earned}, are not taken into account using just the described approach. 

\comment{But the description of the Rule-based AI is very short and they don’t show any examples of rules discovered. -> ja haben keinen Platz. Vlt für die Präsentation nett? @Hayyan}
For that matter, we used the combination of ASP and EC such that e.g. defaults can model recommendations.
The first main issues concerning modeling study requirements in general and using EC was translating examination regulations given in natural languages into formal languages.
We encountered the following problems and challenges:
\begin{itemize}
    \item There are rules that are understood by any human and thus not written.
    \item There is a lot of human interference that allows for exceptions. Exceptions in study plans are not rare.
    \item There are older versions of the examination regulations, which certain students still follow.
\end{itemize}

The second problem we encountered with EC is that almost all events contribute to a single result (e.g. CPs), instead of a majority of events, each initiating new kinds of results. EC is designed for the latter, but in study plans the former holds.
We thus adjusted the EC.
One modification was to differentiate between events that happened and events that are planned.
For planning in the future, one needs to follow the rules.
For events in the past, a record is sufficient and there is no need for further requirement checking.
This allows to add exceptions that are actually inconsistent with the examination regulations.
It was also important to keep track of certain relevant numbers a student has at any point in time, in order to be able to do requirement checking.
This was achieved through results, which events can contribute to.
\textit{Mathematics 1}, for example, adds 9 units to the result \textit{credit points}, after the event of success at it.
A requirement on CPs should consider the general number of CPs collected or just within a field or a time frame. 
For that matter we created the notion of a \textit{result requirement}, which makes sure that the sum of results caused by a subset of events is less than, greater than, or equal to some value.
With all of this in mind, we separated the required rules into three categories:
\begin{itemize}
    \item \textit{Invariant}: Rules about the requirements and the EC modified axiom system.
    \item \textit{Variant by Admins}: Rules about modules and their availability.
    \item \textit{Variant by Student}: Rules about the plan of the student.
\end{itemize}

After that, we were able to translate the examination regulations, module catalogs, and student event logs into rules.
This enables us to perform model as well as conformance checking.

\section{Future steps}\label{sec:fw}
\comment{If the only result we can obtain from conformance checking is which courses are taken in the semester they are supposed to, I do not think we are taking full advantage of those algorithms. Applying a simple query to the database will provide the same results. -> Future work}
Until now, the data are limited to information about exams and is exclusively derived from the CMS.
In a next step, course enrollments will be added to further analyze study behavior of students.
This additional information will give more concrete insights about the students' intended study plan, since at many universities, course enrollments are not automatically coupled to exam registrations. 
While students might start to take a course in the intended semester, thus enroll in it, they might realize that the workload is too high or exam qualification requirements are not fulfilled and refrain from registering for the exam in the end. 
This may also be valuable information considering the instructors' workload as more course enrollments indicate more work during the course and may require larger lecture halls or additional support staff.
As such, this workload needs to be balanced out when planning courses for upcoming semesters

The information stored in the LMS contains valuable information to understand students' learning behavior, as shown in related work.
When combined with activities in the CMS, a more complete view on students' behavior and more direct feedback about the success of the intended plan can be provided.
This feedback can then be used in BuddyAnalytics to help study program designers in improving curricula and recommended study plans, as well as give more informed suggestions for individual study plans.
Possibly, in StudyBuddy, students might be informed about their behavior deviating from a recommended plan and are presented with suggestions suitable to their individual circumstances.

On the theoretical side, the possibilities of a combination of AI and PM are further explored and implemented.
The main focus will be to improve the conformance checking results.
Also, PM conformance checking possibilities will be further explored.
One planned aspect is the extraction of constraints from event logs directly.
We expect to learn rules that are not intended but are beneficial, e.g., \textit{Statistics} is a good preparation for \textit{Introduction to Data Science} and when taken in order, the grade and success rate of the latter improves.
Those rules could be added to the examination regulations rules as defaults.

\section{Conclusion}\label{sec:conclusion}
The AIStudyBuddy project will combine different existing AI and PM frameworks and extend them with new features, making use of the already existing data at universities, to help students and study program designers make more informed decisions about study paths and curricula.
\comment{The authors claim “the first results are promising”, but fail to provide enough results so one can agree with that claim. -> changed}
The first results get positive feedback from students and study program designers.
Currently, only a small fraction of available CMS data was used to produce these results, leaving a lot of potential for future steps.
PM techniques already give valuable new insights to the study program designers and the combination of AI and PM for conformance checking in particular helps overcome restrictions due to the data and rule properties.
Having requirements and recommendations, credit point boundaries, and long-term relations between courses should be included in the system to model examination regulations in a more accurate manner.

\subsubsection{Acknowledgements:} The authors gratefully acknowledge the funding by the Federal Ministry of Education and Research (BMBF) for the joint project AI\-StudyBuddy (grant~no.~16DHBKI016).
\subsubsection{Printversion}
This paper is a postprint version. The published version is \copyright Springer (DOI pending).
%
%
%


\bibliographystyle{splncs04}
\bibliography{mybibliography}

\end{document}